\documentclass[preprint,12pt]{elsarticle}

\usepackage[normalem]{ulem}
\usepackage[usenames,dvipsnames,table,xcdraw]{xcolor}

\usepackage{graphicx}
\usepackage{amssymb}

\usepackage{verbatim}
\usepackage{acronym} 
\usepackage{svg}
\usepackage{amsmath}

\usepackage[colorinlistoftodos]{todonotes}
\usepackage{hyperref}
\usepackage{siunitx}
\usepackage{pgfplots}
\usepackage{pgfgantt}
\usepackage{multirow}
\usepackage{booktabs}

\usepackage[table,xcdraw]{xcolor}

\acrodef{PWM}[PWM]{Pulse Width Modulation}
\acrodef{PFM}[PFM]{Pulse Frequency Modulation}
\acrodef{FPGA}[FPGA]{Field Programmable Gate Array}
\acrodef{PID}[PID]{Proportional Integral and Derivative}
\acrodef{DVS}[DVS]{Dynamic Vision Sensor}
\acrodef{VLSI}[VLSI] {Very Large Scale Integration}
\acrodef{AER}[AER]{Address Event Representation}
\acrodef{PCB}[PCB]{Printed Circuit Board}
\acrodef{FSM}[FSM]{Finite State Machine}
\acrodef{LUT}[LUT]{Look-Up Table}
\acrodef{CPG}[CPG] {Central Pattern Generator}
\acrodef{sCPG}[sCPG] {Spiking Central Pattern Generator}
\acrodef{SNN}[SNN]{Spiking Neural Network}
\acrodef{DOF}[DOF]{degrees of freedom}
\acrodef{HDL}[HDL]{Hardware Description Language}
\acrodef{FW}[FW]{Forward}
\acrodef{BW}[BW]{Backward}

\journal{Neurocomputing}

\begin{document}

\begin{frontmatter}


\title{NeuroPod: a real-time neuromorphic spiking CPG applied to robotics}



\author{Daniel Gutierrez-Galan\fnref{label1}}
\ead{dgutierrez@atc.us.es}

\author{Juan P. Dominguez-Morales\fnref{label1}}
\author{Fernando Perez-Pe\~{n}a\fnref{label2}}
\author{Alejandro Linares-Barranco\fnref{label1}}

\address[label1]{Robotics and Computer Technology Lab. Universidad de Sevilla, Spain.}
\address[label2]{Department of Computer Architecture and Technology, Universidad de C\'{a}diz, Spain.}

\begin{abstract}
Initially, robots were developed with the aim of making our life easier, carrying out repetitive or dangerous tasks for humans. Although they were able to perform these tasks, the latest generation of robots are being designed to take a step further, by performing more complex tasks that have been carried out by smart animals or humans up to date. To this end, inspiration needs to be taken from biological examples. For instance, insects are able to optimally solve complex environment navigation problems, and many researchers have started to mimic how these insects behave. Recent interest in neuromorphic engineering has motivated us to present a real-time, neuromorphic, spike-based Central Pattern Generator of application in neurorobotics, using an arthropod-like robot. A Spiking Neural Network was designed and implemented on SpiNNaker. The network models a complex, online-change capable Central Pattern Generator which generates three gaits for a hexapod robot locomotion. Reconfigurable hardware was used to manage both the motors of the robot and the real-time communication interface with the Spiking Neural Networks. Real-time measurements confirm the simulation results, and locomotion tests show that NeuroPod can perform the gaits without any balance loss or added delay. 

\end{abstract}

\begin{keyword}
Neurorobotics \sep SpiNNaker \sep Central Pattern Generator \sep Spiking Neural Network \sep Neuromorphic Hardware \sep FPGA

\end{keyword}

\end{frontmatter}


\section{Introduction}
\label{S:Intro}
A \ac{CPG} is a neural structure located at a spinal cord level. It can generate rhythm patterns which might be used for movements, such as the generation of various gaits, or swimming \cite{GRILLNER2008}. There is proven evidence of such structures in small animals \cite{Duysens1998} and possibly in humans \cite{Guertin2009,minassian2017}. The activity of these structures is released and mediated by the brain stem and other sub-cortical regions of the brain. Regarding the feedback, the \ac{CPG} receives sensory information to adapt its output to the environment.

Locomotion is probably one of the most complex tasks to be developed by roboticists due to stability issues when several legs are involved \cite{Schilling2013}. Therefore, from a neurorobotics point of view, the idea of these \acp{CPG} is borrowed from biology to implement locomotion in small robots with several legs. The reason for this is that these structures can generate a very stable pattern even without sensory information or brain activity. In fact, some cats that suffered severe spinal cord injuries, recovered their gaits after treadmill training sessions \cite{Vogelstein2008}. 
In this paper, we borrow this \ac{CPG} feature: the ability to generate patterns in an open-loop manner. This can be useful as a first approach for the use of \acp{CPG} in neurorobotics, which is the target field of this paper.   

There are many works where a \ac{CPG} has been used within robotics; some of them mimic the idea of a \ac{CPG}, although without implementing a spiking neural network. Instead, they modelled the \ac{CPG} using differential equations of coupled oscillators. Examples are: \cite{Sartoretti2018}, where the authors used a hexapod robot and they included feedback, \cite{barron2010}, where the Van der Pol oscillator model was used and implemented on a \ac{FPGA}, and \cite{Crespi2008}, where the authors used a swimming and crawling fish robot and implemented the \ac{CPG} on a microcontroller by solving the equations of coupled oscillators. 

More closely related works, where neuromorphic hardware was used or a \ac{SNN} was proposed, are: \cite{still2001}, where the authors developed an analog neuromorphic dedicated chip which allocates coupled oscillators and a learning procedure to have the desired output, although they did not use well-known neuron models, and \cite{Donati2014}, in which the authors designed and implemented several \acp{CPG} segments to drive a lamprey-like robot. The \acp{CPG} were implemented using neuromorphic hardware in \cite{qiao2015}, although online changes of the pattern generated by the \ac{CPG} are not provided. Likewise, the work presented in \cite{Brayan_2017} proposed the implementation of the \ac{CPG} using a \ac{SNN} implemented on SpiNNaker \cite{Furber2014a}, although it does not offer real time nor online change of the pattern produced by the \ac{CPG}.

The work presented in this paper is based on the one presented in \cite{Brayan_2017}. The objective is to implement a \ac{CPG} closely related to its biological counterpart including plausible biological features in SpiNNaker and ready to be used within robotics: a hexapod robot was used to validate the design. The novelties of this research are: real-time operation of the \ac{CPG} and online reconfiguration of the gait produced by the \ac{CPG}.  

This paper is structured as follows: subsection \ref{S:mat} describes the materials used and subsection \ref{subsect_Methods} describes how the research was conducted and all the details needed to replicate the experiments. Then, the results achieved by the implementation of the \ac{CPG} in a small robotic platform using neuromorphic hardware are described in section \ref{S:Results}. Finally, section \ref{S:DiscussionAndConclusions} presents a discussion over the obtained results and the conclusions drawn from this study.

\section{Materials and Methods}
This section describes the materials that were used in this project to design, assemble, and control this neuromorphic robot, as well as the methods applied to obtain the results shown in section \ref{S:Results}.
\subsection{Materials}
\label{S:mat}

\begin{figure}[ht]
\centering\includegraphics[width=1\linewidth]{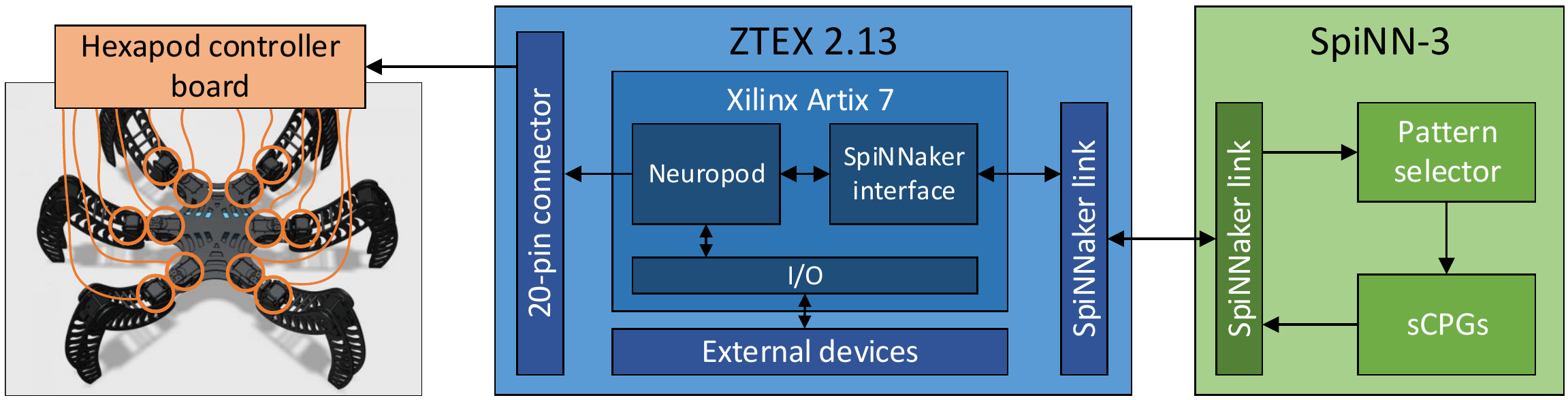}
\caption{Block diagram of the entire system. It is composed of the SpiNNaker board, an FPGA-based board, and a 3D-printed hexapod robot frame.}
\label{fig_BD_system}
\end{figure}

The NeuroPod robot is divided into three main parts, and each of these has a specific role or functionality. These parts are the \ac{CPG}, designed using a \ac{SNN} and implemented on a neuromorphic hardware platform. This CPG generates the gait patterns. The movement controller takes the movement information from the \ac{CPG} and controls a set of servomotors through an FPGA-based board. Finally, the skeleton defines the shape of the robot and also performs the movements. Further details are provided in the following sections. Figure \ref{fig_BD_system} shows a global overview of the NeuroPod as a block diagram; it also shows the main parts and how they interface with each other.

\subsubsection{Robotic platform}

An hexapod robot is a six-legged robot inspired by arthropod insects, such as ants or flies, among others. According to biology, the body of these insects can be divided into three different regions: the head, the thorax and the abdomen. Moreover, each part could be sub-divided in segments according to their features.

The head is composed of eyes (located in the ocular segment) and a pair of antennae. Both the eyes and the antennae are used to collect sensory information about the environment, allowing the movement of the insect in complex scenarios by performing an obstacle avoidance task \cite{douglass1995visual}\cite{milde2015bioinspired}. The thorax is composed of three segments: the prothorax, the mesothorax and the metathorax. Each segment has a pair of legs, and there are six in total. Up to five parts can be identified in each leg, although only three of these parts are relevant to motion: coxa, femur and tibia. Finally, the abdomen contains the vital organs of the insect, such as the respiratory or reproductive systems.

Recent focus on the development of smart robots by mimicking biological processes has motivated many research groups to develop accurate models of hexapod insects. HECTOR \cite{schneider2012hector} is an example of that, where both the body features and the movements were inspired by the morphological details of the stick insect \textit{Carausius morosus}.

In this work, a 3D-printed hexapod robot was used, based on the model featured in \cite{Brayan_2017}. The original design\footnote{https://www.thingiverse.com/thing:1021540 (checked on April'2019)} was adapted by designing a new body frame to allocate the electronic devices on it. The frame dimensions are 20 x 89 x 90 mm (height, width, depth), without the legs.

\begin{figure}[t]
\centering\includegraphics[width=0.95\linewidth]{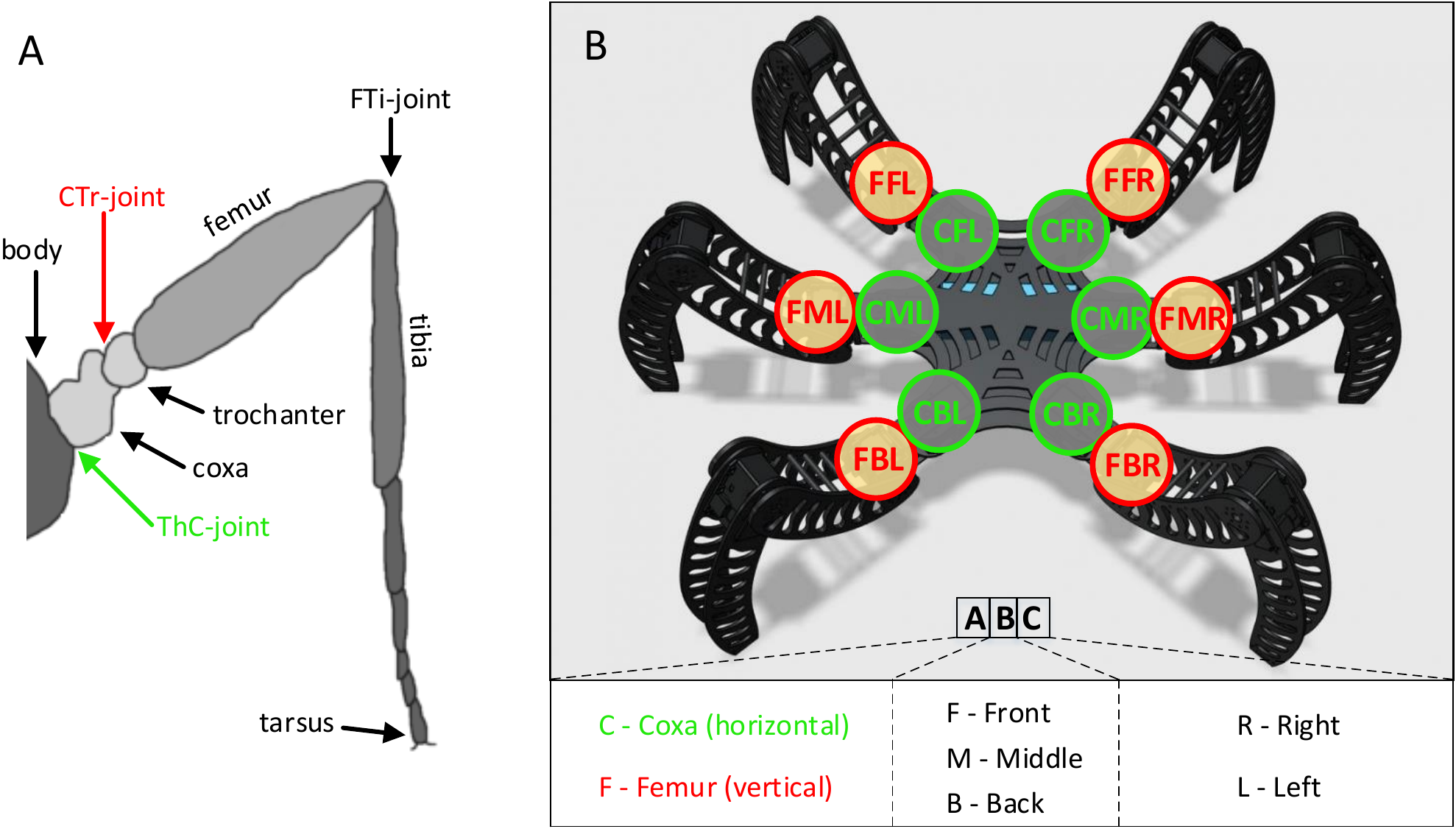}
\caption{A) Biological representation of an arthropod's leg anatomy. B) Hexapod robot leg actuator IDs.}
\label{fig_hexapod_overview}
\end{figure}

According to \cite{buschges2008organizing}, insect legs are defined as multi-segmented limbs. Each leg consists of more than 5 segments, as it is represented in Fig.\ref{fig_hexapod_overview}A. However, only three of them are used when performing a movement: the coxa, the femur and the tibia. This is due to the fact that three main leg joints can be found in an insect leg: the thoraco-coxal (ThC-) joint, the coxa-trochanteral (CTr-) joint and the femur-tibia (FTi-) joint. The ThC-joint is responsible for carrying out back and forth movements (horizontal axis), the CTr-joint enables elevation and depression and the FTi-joint allows extension and flexion (both in the vertical axis).

The leg of each hexapod has three \ac{DOF}, one per joint. However, to develop NeuroPod we only considered two of them because the movement of the robot can be performed mainly using the coxa and the femur \cite{rostro2015cpg}. Thus, only twelve \ac{DOF} were taken into account, instead of eighteen, to implement the gait patterns.


In order to provide motion, one servomotor was placed on each joint, making a total of twelve servomotors (Ref. SG90). The maximum rotation angle is 180 degrees, although this range could be reduced due to mechanical constraints of the body design and the position of the servo on it. 

Therefore, a calibration is required. After the calibration process, and knowing that the theoretical operation speed of the selected servos is 0.12 s/60 degrees, we were able to estimate the pattern period, which can be defined as the minimum time the robot needs to reach the backward position, starting from the forward position, and then reach the forward position again. Measurements of these pattern periods are presented in section \ref{S:Results}.

\subsubsection{SpiNNaker}

The SpiNNaker project is based on a massive parallel multicore computing system that is able to run very large \acp{SNN} in real time \cite{furber2013overview}. The architecture of the SpiNNaker chip, which has an asynchronous packet switching network, makes it very efficient for neuromorphic applications \cite{plana2007gals}.

\begin{figure}[ht]
\centering\includegraphics[width=1\linewidth]{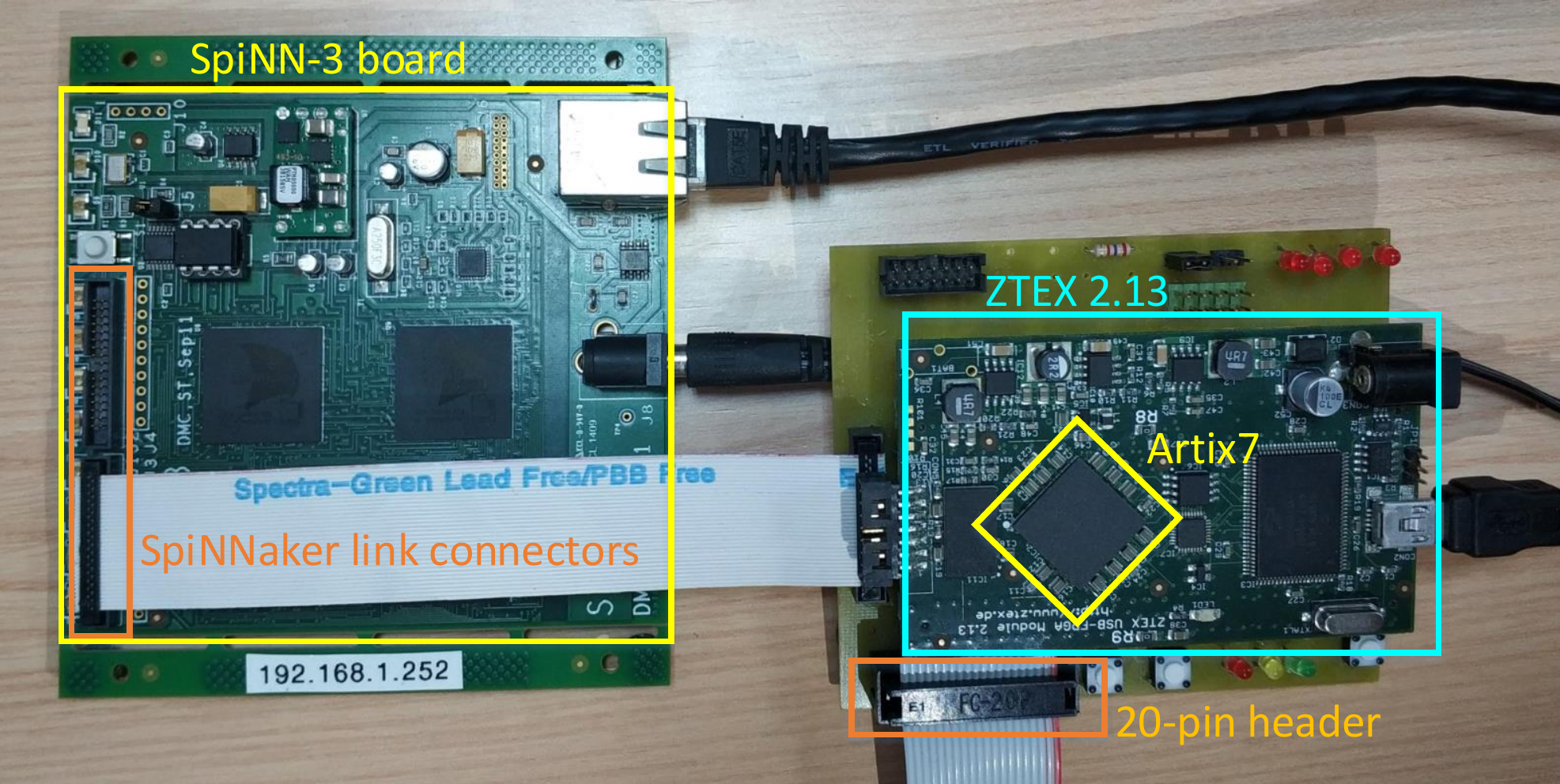}
\caption{SpiNN-3 machine and ZTEX 2.13 board.}
\label{fig_SpiNN3}
\end{figure}

In this work, the SpiNN-3 machine (4 SpiNNaker chips, 72 200MHz ARM9 cores) was used to implement the \ac{SNN} model, which is described in section \ref{subsect_Methods}. The device is shown in Fig. \ref{fig_SpiNN3}. This board has an interface, 100 Mbps Ethernet link, which is used to control the SpiNNaker machine from the computer. It also has two spinn-link connectors that enable a connection to external devices such as \acp{FPGA} and neuromorphic sensors: retinas or cochleas. This board was connected to an \ac{FPGA} for real-time input/output communication.

\subsubsection{Reconfigurable hardware board} \label{subsubsec_ReconfigurableHardware}

An \ac{FPGA}-based board was used to implement a digital system design, which controls the neuromorphic robot platform. This approach was considered in other similar works to implement a hardware version of \acp{CPG}, such as \cite{barron2010} and \cite{rostro2015cpg}, and also in the field of neurorobotics and neuromorphic engineering\cite{yousefzadeh2017multiple}. This reconfigurable hardware offers flexibility against analog designs and adaptability in real time, in case of system failures.

From the Xilinx Artix-7 family, the XC7A75T chip was used, mounted on the ZTEX 2.13 USB-FPGA board with a 48 MHz clock source. This FPGA chip offers around 75500 logic cells, 100 GPIOs, USB 2.0 interface and DDR3 SDRAM memory. This board serves as a daughter board located on a custom base-board provided with several components: LEDs, user buttons, an AER 20-pin interface and a SpiNNaker link interface. Those interfaces will be used by the SpiNNaker machine to manage the hexapod robot through the FPGA board. An extended explanation is provided in section \ref{subsubsect_Platform}.

\subsection{Methods} \label{subsect_Methods}

\subsubsection{\ac{CPG}}
A \ac{CPG} is a neural network in which interconnected excitatory and inhibitory neurons produce an oscillatory, rhythmic output as a motor pattern, such as walking, flying, running or swimming, with the absence of rhythmic inputs. In this work, we focus on three specific gaits: walk, trot and run, which are selected based on previous working bio-inspired implementations for hexapods \citep{rostro2015cpg, Brayan_2017}.

\begin{figure}[!t]

\centering\includegraphics[width=1\linewidth]{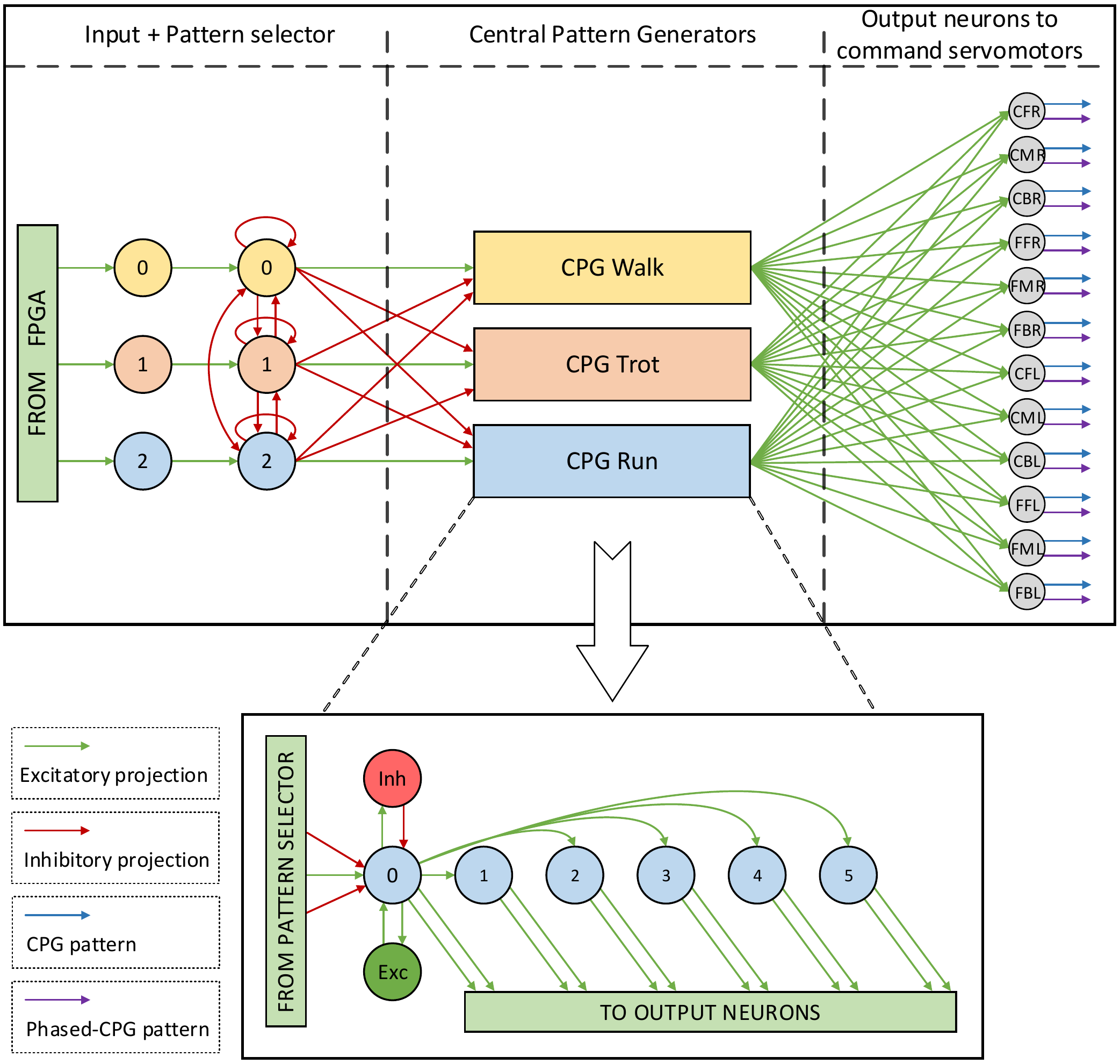}
\caption{Diagram of the spiking neural network model used (top) with an in-depth view of the CPG architecture (bottom).}
\label{fig_CPGSNN}
\end{figure}

Fig. \ref{fig_CPGSNN} (bottom) shows the basic structure for each of the \acp{CPG} implemented in this work. It consists of eight neurons: two neurons for the \ac{sCPG} and six output neurons to command the servomotors. The green and red neurons (Fig. \ref{fig_CPGSNN}) make the other neurons, ranging from 0 to 5, fire within different timings generating the selected gait. Each of these six neurons are then connected to two output neurons which will command two different servomotors. This is achievable due to the symmetry of the robot: pairs of servomotors always perform the chosen gait, independently of the selected \ac{sCPG}.

Three \acp{sCPG} following this basic architecture (one per gait) are enclosed within a global \ac{SNN} model shown in Fig. \ref{fig_CPGSNN} (top). This global network acts as a mechanism to select which of the \acp{sCPG} has to be enabled in order to start generating the gait, inhibiting the other two at the same time. It receives a single spike from the \ac{FPGA} with a specific neuron address (0, 1 or 2) (the \ac{AER} protocol is used) indicating the gait pattern that needs to be generated. When this spike reaches the pattern selector population (the three neurons that are closest to the \acp{sCPG}), this group of neurons transmit this spike to the correct \ac{sCPG}, while inhibiting the other two. This mechanism allows real-time gait changes by activating the appropriate \acp{sCPG} without introducing a long delay (hundreds of ms), which is really important for real-time robotics applications.

A single spike is needed by the selected \ac{sCPG} to start generating the spiking pattern. The spikes fired by the six \ac{sCPG} neurons are sent to the last layer of the model, which consists of twelve neurons corresponding to each of the hexapod servomotors. These spikes are transmitted back to the \ac{FPGA}, where a circuit commands each of the servomotors using the live output spikes.

As is shown in Fig. \ref{fig_CPGSNN}, each neuron of the output layer, has two outputs from SpiNNaker to the \ac{FPGA}. The first one is the regular spiking pattern generated by the \ac{sCPG} to command the servomotors (extension action). The second one is exactly the same pattern, but phased with a delay of 1 tick, needed by the \ac{FPGA} to command the servomotors back to the standard position (performing the flexion action). 

\subsubsection{Digital system} \label{subsubsect_Platform}

As previously mentioned in section \ref{subsubsec_ReconfigurableHardware}, a digital system is needed to implement the neuromorphic robotic controller. This task is often carried out by using an \ac{FPGA}-based board, running a custom digital system. Designs are generally implemented using \ac{HDL}, which allows defining any digital circuit model by describing either its behavior or its components' interconnection. Each component of the design is also known as a module, and many functional modules can be encapsulated by a top module, which defines both the input and output signals of the digital circuit.

\begin{figure}[ht]
\centering\includegraphics[width=1\linewidth]{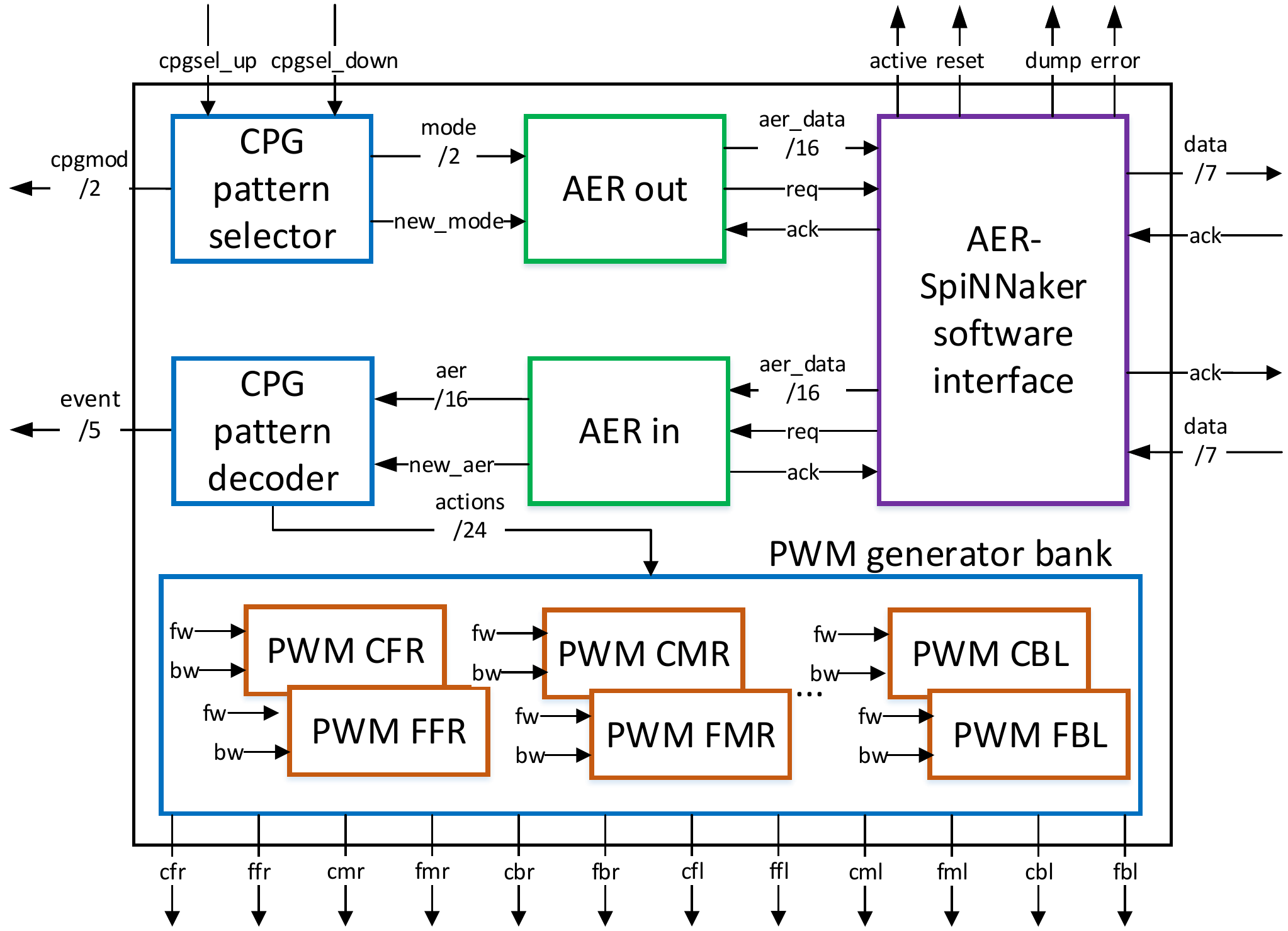}
\caption{NeuroPod FPGA top module overview.}
\label{fig_NeuroPod_module}
\end{figure}

Fig. \ref{fig_NeuroPod_module} shows an overview of the proposed implementation of the NeuroPod control system top module. It performs three main functions: to select the gait which the \ac{sCPG} implemented on SpiNNaker will generate, to send the gait information to the SpiNNaker machine, receive the live output pattern from it and, finally, to generate the \ac{PWM} signals to control the servomotors. Further details are given next, starting with the pattern selector and then following accordingly with the work flow.

First, the \ac{CPG} pattern selector module was implemented as a 2-bit up/down unsigned counter. Both up and down signals are declared as input and they are directly mapped to two buttons located on the base board in which the ZTEX board is connected. The current counter value indicates the gait: 0 for walking, 1 for trotting and 2 for running. This information is shown to the user by means of a pair of LEDs in binary format.

Every time the CPG pattern selector changes its value, an interruption is generated through the signal \textit{new\_mode} to notify that there is a new data available to the next module. This component is the \ac{AER} out module, which converts a 2-bit value to a 16-bit \ac{AER} event, and also handles the \ac{AER} handshake protocol (\textit{REQUEST} and \textit{ACK} signals). This conversion to \ac{AER} is carried out since the system needs to send information to the SpiNNaker to generate the pattern. 

The SpiNNaker link interface is based on the 2-of-7 protocol. Then, since most of the neuromorphic sensors use the AER protocol, to allow the communication in real-time between AER devices and the SpiNNaker, an \ac{AER}-SpiNN \ac{HDL} module was developed by the SpiNNaker team \cite{plana_2014}.

It takes \ac{AER} events as input following the \ac{AER} protocol, and generates packets under the 2-of-7 protocol for the communication from the \ac{AER} device to the SpiNNaker board. For the reverse communication, it takes 2-of-7 packets and generates \ac{AER} events. In addition, this module provides four status signals to check in real-time if the communication is working properly.

\ac{AER} events received by the SpiNNaker, generated by the \ac{AER}-SpiNN module, are captured by the \ac{AER} in module, which implements the handshake and sets the value of the event as a 16-bit output signal. In the same way as the \ac{AER} out module, an interruption is enabled every time the \ac{AER} in module receives a new input event.

Those input events correspond to the spikes fired by the output layer neurons of the \ac{SNN} implemented on SpiNNaker. Since each output neuron manages the position of one servomotor, these events have to be mapped to the correct one. Therefore, depending on the address of the event, an action is performed over a servomotor. 

There are two actions available: either move the servomotor towards the forward position or move the servomotor towards the backward position. A decoding scheme is summarized in Table \ref{Decodification_scheme_table}, where each column represents the joints of the NeuroPod following the same nomenclature as in Fig. \ref{fig_hexapod_overview}: the FW row means forward action, the BW row means backward action, and each value is the \ac{AER} event which triggers the action. 


The commands received are converted to motion through a \ac{PWM} generator block, which receives the decoded \ac{AER} events by means of a 24-bit signal (one enabling signal per action). This \ac{PWM} generator block was implemented instantiating as many \ac{PWM} generators as the number of joints there are in the NeuroPod.

\begin{table}[!h]
\centering
\caption{AER decodification scheme.}
\label{Decodification_scheme_table}
\resizebox{\textwidth}{!}{%
\begin{tabular}{c|c|c|c|c|c|c|c|c|c|c|c|c|}
\cline{2-13}
 & \textbf{CFR} & \textbf{FFR} & \textbf{CMR} & \textbf{FMR} & \textbf{CBR} & \textbf{FBR} & \textbf{CFL} & \textbf{FFL} & \textbf{CML} & \textbf{FML} & \textbf{CBL} & \textbf{FBL} \\ \hline
\multicolumn{1}{|c|}{\textbf{FW}} & 0 & 1 & 2 & 3 & 4 & 5 & 6 & 7 & 8 & 9 & 10 & 11 \\ \hline
\multicolumn{1}{|c|}{\textbf{BW}} & 12 & 13 & 14 & 15 & 16 & 17 & 18 & 19 & 20 & 21 & 22 & 23 \\ \hline
\end{tabular}%
}
\end{table}


The \ac{PWM} generator implemented includes some features that make the NeuroPod motion control easier: up to three pulse width values can be set in the same VHDL module instead of only one. These values were used to define the positions that the servomotor had to reach. Those positions are: forward, backward (corresponding to the actions) and home.
Two control signals were added to the \ac{PWM} generator to select the configuration of the module: \textit{fw}, which enables the generation of the \ac{PWM} signal associated to the FW position, and \textit{bw}, which enables the generation of the \ac{PWM} signal associated to the BW position. 

This module has two input signals, which are connected following the scheme shown in Table \ref{Decodification_scheme_table}. When a control signal is set to high, either through a single pulse or constant signal, the pulse width value associated to that control signal is loaded in the configuration register. Then, the \ac{PWM} output signal changes automatically, moving the servomotor to the commanded position. That output signal is held until the module receives a different action command. Finally, the home position is only activated when the global reset signal is released.

\section{Results}
\label{S:Results}

\begin{figure}[ht]
\centering\includegraphics[width=0.85\linewidth]{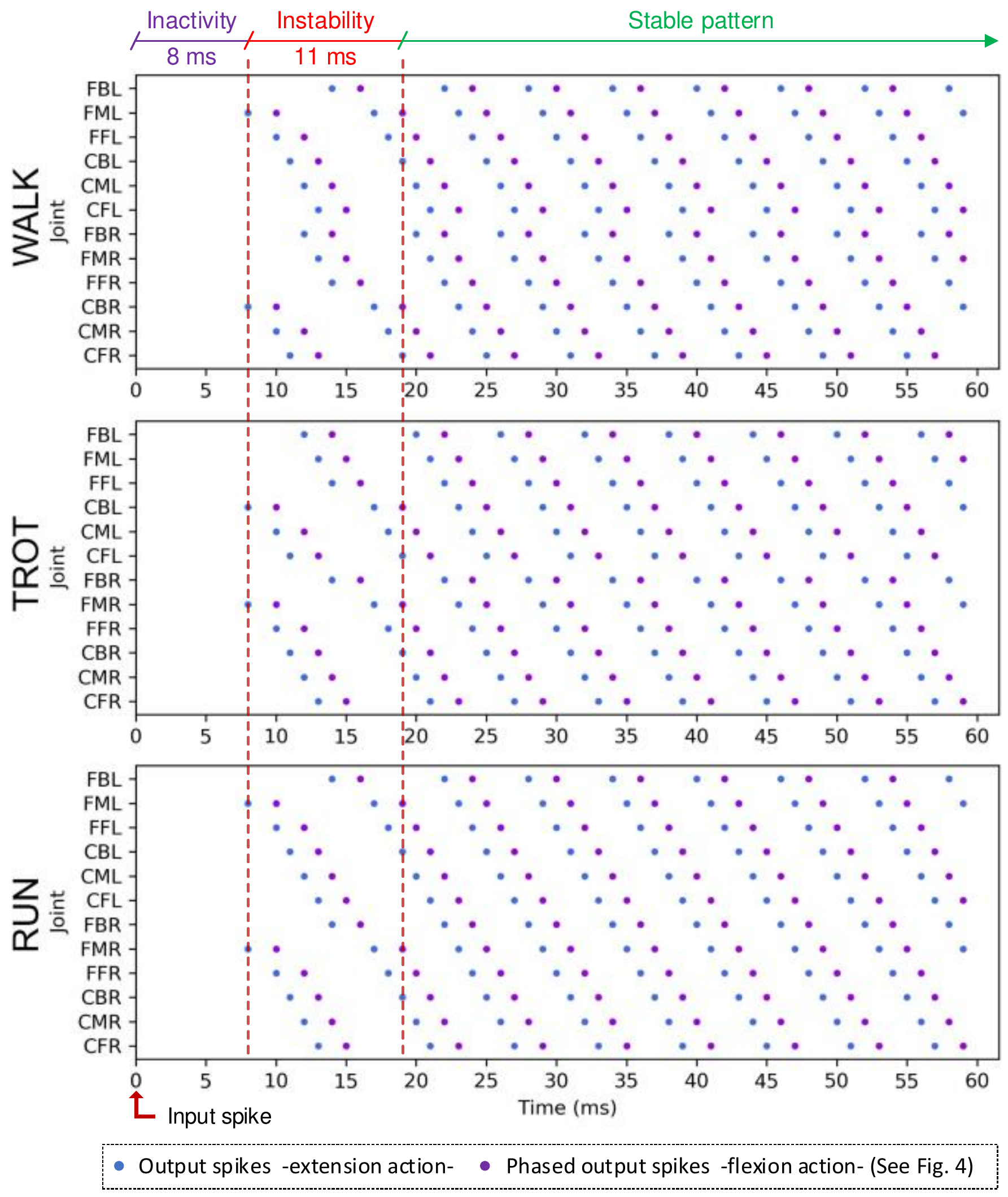}
\caption{Output spikes for each gait pattern simulated on SpiNNaker.}
\label{fig_outputSpikes_mix}
\end{figure}

The results obtained for the simulation of each of the gait patterns on SpiNNaker are shown in Fig. \ref{fig_outputSpikes_mix}. This figure shows the output spikes that the \ac{sCPG} generates.

\begin{figure}[ht]
\centering\includegraphics[width=1\linewidth]{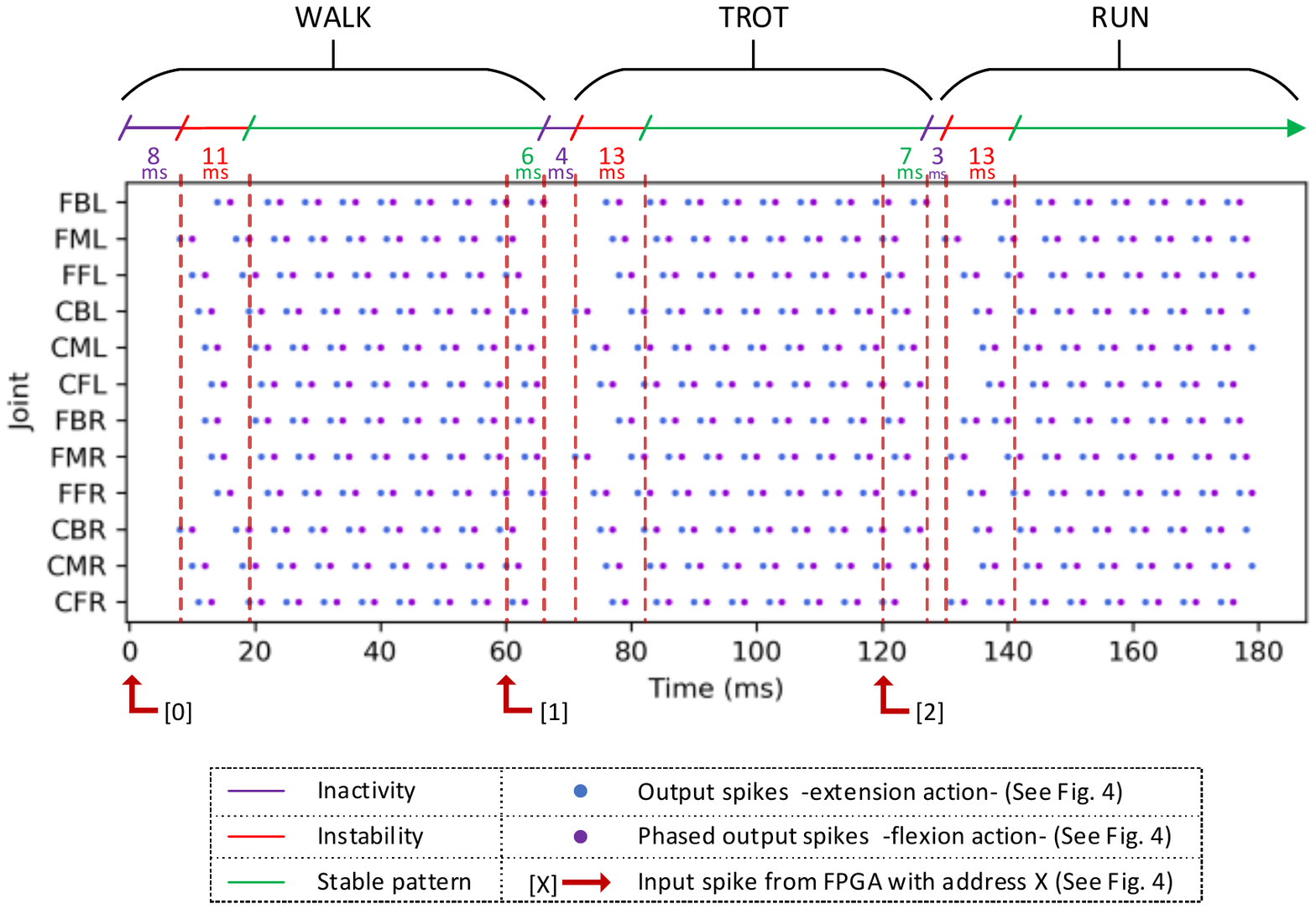}
\caption{Output spikes from the SpiNNaker simulations that show the gait pattern change behavior.}
\label{fig_outputSpikes_change}
\end{figure}

Then, in Fig. \ref{fig_outputSpikes_change}, the same plot is shown for a different scenario in which we simulated and tested the behavior of the SNN when forcing the system to change from a specific sCPG to a different one. The figure shows how the system is able to change from walk to trot and then to run, generating a stable pattern for each gait after a specific period of time, which, in this case, is 23 milliseconds. This delay is the time that the network takes to inhibit the neurons related to the previous gait that was being executed plus the time that the neurons related to the current pattern take to start generating the correct firing output in a stable way. Different delays related to these simulations were measured and are presented in the image.

It is important to mention that these delays were measured in simulation. For a real-time scenario, the delays do not match these values since, due to the fact that the servomotors were not able to work at such speed, we had to set the time\_scale\_factor parameter on the SpiNNaker board to 100. This made the whole simulation run 100 times slower in real time, which is very convenient for this approach.

 
Regarding the \ac{FPGA}, a study of the VHDL module was carried out. A post-implementation resources consumption report was generated, and also post-implementation simulations were performed to measure the delays of every single VHDL module. The obtained results from those analysis are shown in Table \ref{table_fpga_results}. To obtain these times, a clock source of 48MHz was used.

\begin{table}[t]
\centering
\caption{FPGA resources consumption and delays}
\label{table_fpga_results}
\resizebox{\textwidth}{!}{%
\begin{tabular}{c|c|c|c|}
\cline{2-4}
 & \multicolumn{2}{c}{\cellcolor[HTML]{C0C0C0}\textbf{Resources consumption}} & \cellcolor[HTML]{C0C0C0}\textbf{Delays} \\ \cline{2-4} 
\multirow{-2}{*}{} & LUTs & Registers & Time (Clock cycles) \\ \hline
\rowcolor[HTML]{EFEFEF} 
\multicolumn{1}{|c||}{\cellcolor[HTML]{EFEFEF}\begin{tabular}[c]{@{}c@{}}CPG pattern\\ selector\end{tabular}} & 2 (\textless{}0.01\%) & 4 (\textless{}0.01\%) & 20.83 ns (1) \\ \hline
\multicolumn{1}{|c||}{\begin{tabular}[c]{@{}c@{}}AER \\ out\end{tabular}} & 7 (0.01\%) & 5 (\textless{}0.01\%) & 104.15 ns (5) \\ \hline
\rowcolor[HTML]{EFEFEF} 
\multicolumn{1}{|c||}{\cellcolor[HTML]{EFEFEF}\begin{tabular}[c]{@{}c@{}}AER-SpiNN\\ interface\end{tabular}} & 213 (0.45\%) & 272 (0.29\%) & 1374.78 ns (66) \\ \hline
\multicolumn{1}{|c||}{\begin{tabular}[c]{@{}c@{}}AER \\ in\end{tabular}} & 7 (0.01\%) & 10 (0.01\%) & 41.66 ns (2) \\ \hline
\rowcolor[HTML]{EFEFEF} 
\multicolumn{1}{|c||}{\cellcolor[HTML]{EFEFEF}\begin{tabular}[c]{@{}c@{}}CPG pattern\\ decoder\end{tabular}} & 12 (0.03\%) & 24 (0.03\%) & 20.83 ns (1) \\ \hline
\multicolumn{1}{|c||}{\begin{tabular}[c]{@{}c@{}}PWM generator\\ block\end{tabular}} & 720 (1.53\%) & 576 (0.61\%) & 1895.53 ns (91) \\ \hline \hline
\rowcolor[HTML]{C0C0C0} 
\multicolumn{1}{|c||}{\cellcolor[HTML]{C0C0C0}\textbf{\begin{tabular}[c]{@{}c@{}}NeuroPod\\ Top\end{tabular}}} & \textbf{986 (2.09\%)} & \textbf{893 (0.95\%)} & \textbf{3457.78 ns (166)} \\ \hline
\end{tabular}%
}
\end{table}

The amount of resources used by the top module is around 2.1 \% of the available LUTs and around 1\% of the available number of registers. This low resources consumption allows implementing more complex spike-based motor control modules \cite{jimenez2012neuro} as well as improving the NeuroPod top module by including input information about the environment.
In addition, the delay added by the full design, in the worst case, is almost \SI{3.5}{\micro\second}, which is irrelevant compared to the delays presented in Fig. \ref{fig_outputSpikes_change}.


After the simulation results were obtained, a real-time analysis of the full system was carried out.

Both the latency from a high level command to the generation of the \ac{CPG} and the actual motion of the leg were measured using an oscilloscope. These delays can be discarded since they are three orders of magnitude lower than the time taken by the SpiNNaker to generate the \ac{sCPG} which is 800 ms (the simulated time updated with the  time\_scale\_factor). 

According to that latency and the time that a gait cycle takes to be performed, we can conclude that the theoretical maximum value of the movement speed is 1.66 cm/s. 
  
After that, we compared the simulation time with the real-time delays. To this end, four cases of study were defined: resting to moving, stabilization time, movement's period, and change time between two different gaits.

There were time differences since the parameter time\_scale\_factor was set to 100 in the SpiNNaker script in order to slowdown the SpiNNaker output.

\section{Discussion and conclusions}
\label{S:DiscussionAndConclusions} 
A real time \ac{sCPG} using neuromorphic hardware is presented. As was stated in the introduction, several legs, up to six, are controlled. The delays introduced in the open-loop control are very low, in the order of ms (see Table~\ref{table_fpga_results}). Previous works have a 50ms delay of propagation \cite{Donati2014}, or a converge time of 5 seconds in \cite{crespi2006}. Our results show a time of 23 ms (worst case) to converge and approx. 20ms to propagate the gait. 
\\
Another difference with \cite{Donati2014} is that we propose to use SpiNNaker \cite{furber2014spinnaker} instead of the neuromorphic chip ROLLS \cite{qiao2015}. Also, six neurons less than \cite{Donati2014} are used in this work.

As a future work, we propose to use some voice commands and a neuromorphic auditory sensor \cite{jimenez2017binaural} to change the gait. This real-time change will be available based on the classification achieved by an \ac{SNN} connected to the \ac{sCPG}. \\
Furthermore, feedback will be introduced in the \ac{sCPG} by including either touch or elevation sensors in the lower part of the leg of the robot. This feedback will be used to learn in order to provide learning capabilities to the system.

In this work, we presented and achieved what we believe is the first implementation of a real-time neuromorphic spiking \ac{CPG} to command a hexapod robot using SpiNNaker. Furthermore, we included the possibility to change between three different gaits online.

Demonstration video is also available\footnote{\url{https://youtu.be/YZYAPDJHvLI}}.

\section*{Acknowledgements}
This work was supported by the Spanish grant (with support from the European Regional Development Fund) COFNET (TEC2016-77785-P). The work of D. Gutierrez-Galan was supported by a "Formaci\'{o}n de Personal Investigador" Scholarship from the Spanish Ministry of Education, Culture and Sport. This work was carried out during a research internship of D. Gutierrez-Galan and Juan P. Dominguez-Morales in the Department of Computer Architecture and Technology (Universidad de C\'{a}diz, Spain).

\section*{References}
\bibliographystyle{model1-num-names}
\bibliography{references.bib}

\end{document}